\documentclass[lettersize,journal]{IEEEtran}
\usepackage{amsmath,amsfonts}
\usepackage{algorithmic}
\usepackage{algorithm}
\usepackage{array}
\usepackage[caption=false,font=normalsize,labelfont=sf,textfont=sf]{subfig}
\usepackage{textcomp}
\usepackage{stfloats}
\usepackage{url}
\usepackage{verbatim}
\usepackage{graphicx}
\usepackage{cite}
\usepackage{multirow}
\usepackage{color}
\begin{document}

\title{Defect Transformer: An Efficient Hybrid Transformer Architecture for Surface Defect Detection}

\author{Junpu Wang, Guili Xu, Fuju Yan, Jinjin Wang and Zhengsheng Wang
\thanks{This work was supported in part by the National Key Research and Development Plan under Grant 2018YFB2003803, in part by the National Natural Science Foundation of China under Grant 62073161, 61905112 and U1804157, and in part by the Postgraduate Research $\&$ Practice Innovation Program of Jiangsu Province under Grant KYCX21$\_$0202.}

\thanks{Junpu Wang, Guili Xu$^{*}$ and Fuju Yan are with College of Automation Engineering, Nanjing University of Aeronautics and Astronautics, and also with Nondestructive Detection and Monitoring Technology for High Speed Transportation Facilities, Key Laboratory of Ministry of Industry and Information Technology, Nanjing 211106, China. $^{*}$ Corresponding author: Guili Xu (guilixu2002@163.com)}
\thanks{Jinjin Wang is with College of astronautics, Nanjing University of Aeronautics and Astronautics, Nanjing 211106, China.}
\thanks{Zhengsheng Wang is with College of Science, Nanjing University of Aeronautics and Astronautics, Nanjing 211106, China.}
}

\markboth{IEEE TRANSACTIONS ON INSTRUMENTATION AND MEASUREMENT,~Vol.~14, No.~8, August~2021}%
{Shell \MakeLowercase{\textit{et al.}}: A Sample Article Using IEEEtran.cls for IEEE Journals}


\maketitle

\begin{abstract}
Surface defect detection is an extremely crucial step to ensure the quality of industrial products. Nowadays, convolutional neural networks (CNNs) based on encoder-decoder architecture have achieved tremendous success in various defect detection tasks. However, due to the intrinsic locality of convolution, they commonly exhibit a limitation in explicitly modeling long-range interactions, which is critical for pixel-wise defect detection in complex cases, e.g., cluttered background and illegible pseudo-defects. Recent transformers are especially skilled at learning global image dependencies, but with limited local structural information necessary for detailed defect location. To overcome the above limitations, we propose an efficient hybrid transformer architecture for surface defect detection, termed Defect Transformer (DefT), which incorporates CNN and transformer into a unified model to capture local and non-local relationships collaboratively. Specifically, in the encoder module, a convolutional stem block is firstly adopted to retain more detailed spatial information. Then, the patch aggregation blocks are used to generate multi-scale representation with four hierarchies, each of them is followed by a series of DefT blocks, which respectively include a locally position-aware block for local position encoding, a lightweight multi-pooling self-attention to model multi-scale global contextual relationships with good computational efficiency, and a convolutional feed-forward network for feature transformation and further location information learning. Finally, a simple but effective decoder module is devised to gradually recover spatial details from the skip connections in the encoder. Extensive experiments on three datasets demonstrate the superiority and efficiency of our method compared with other CNN- and transformer-based networks.
\end{abstract}


\section{Introduction}
\begin{figure}
  \centering
  \includegraphics[width=0.50\textwidth]{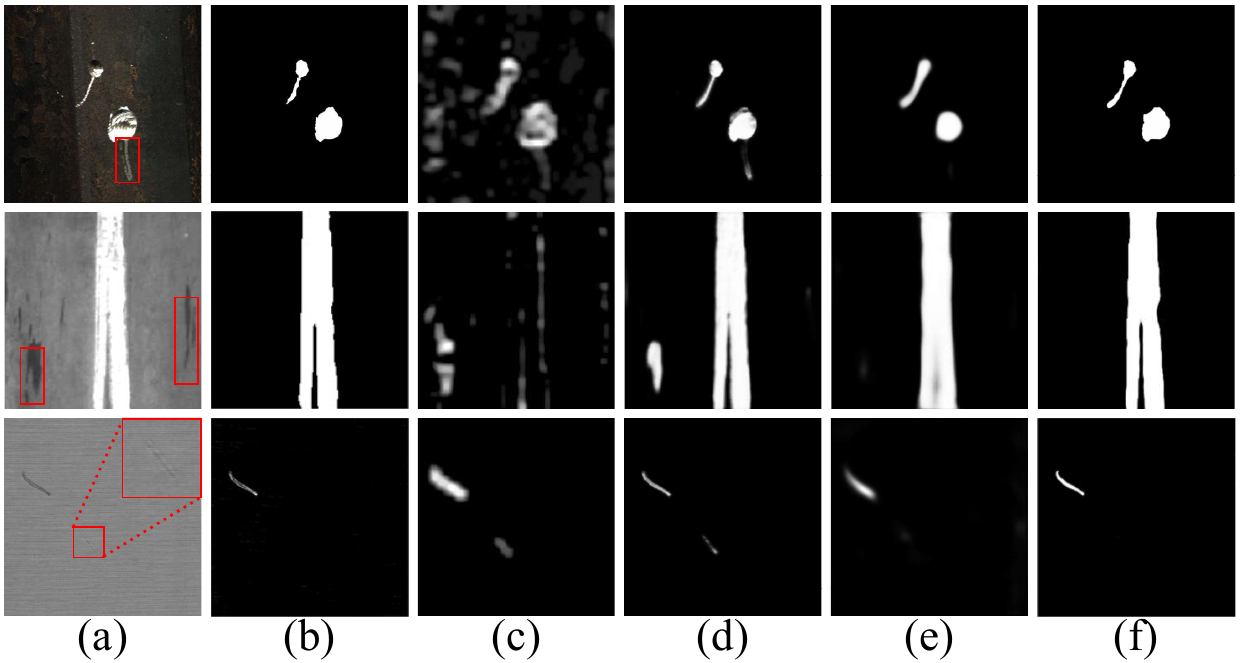}\\
  \caption{Challenging samples with illegible pseudo-defects and cluttered backgrounds in surface defect detection. The red boxes highlight the regions with error-prone pseudo-defects. Column (a) and (b) show the source image and ground truth, followed by different detection methods: (c) KRPCA-NTV [14], (d) UNet [16], (e) SETR [28] and (f) Ours.}\label{fig1}
\end{figure}

\IEEEPARstart{S}{ince}  the instability of industrial equipment and environment, the defects inevitably appear on the product surfaces, thus causing adverse impact on the manufacturing efficiency and profits. Therefore, surface defect detection has become an essential process in the actual production lines. Along with the need of automatic high-precision inspection, computer vision-based methods have evolved into a mainstream to take the place of early manual detection, which encounters the problems of low accuracy and low efficiency. In the past decades, many attempts [1] have been made and justified impressive success, but there is still some room for improvement, especially for exactly distinguishing defects from the background artifacts. As shown in Fig.~\ref{fig1}, there are many illegible rusts, impurities and wrinkles, i.e. so-called pseudo-defects, spreading on the cluttered industrial surface images, they share similar features with genuine defects, making it difficult to give correct recognition even for humans.

Initially, numerous traditional methods based on hand-crafted features and heuristic priors are exploited to recognize defects effectively, which can be roughly divided into three categories: statistical- [2-6], spectral- [7-10], and model-based [11-15] methods. However, these methods generally rely on the accuracy of various assumptions and image priors, which are not always justified in different industrial scenarios. In addition, hand-crafted features mainly explore low-level visual cues, can hardly capture high-level semantic relations between the defects and contexts, leading to insufficient robustness in some challenging cases. As illustrated in Fig.~\ref{fig1}, a representative traditional method KRPCA-NTV [14] produces poor detection results when dealing with error-prone pseudo-defects, yielding a high false-positive rate.

In recent years, the convolutional neural networks (CNNs)-based methods such as UNet [16] have attracted increasing attention in defect detection tasks, attributing to the excellent ability in modeling the low-to-high level features. In general, most CNN-based models attempt to obtain high-level semantic knowledge (large receptive field) to better distinguish salient objects by interleaving multiple 2D convolution-like structures. However, the efficacy of these approaches is confined due to the fact that empirical receptive fields in the CNNs are smaller than the theoretical ones [17]. Therefore, a lot of technologies attend to model more global contextual dependencies for CNNs using large kernels [18][19], atrous convolutions [20][21], attention mechanisms [22-24], et al. Despite being favorable, the intrinsic locality of 2D convolution still limits the access to gather the global context of all pixels in the whole image. As can be seen in Fig.~\ref{fig1}, a typical CNN-based method, UNet [16], achieves better results than traditional method, but still yields misjudgment in some images with extremely complicated background. Hence, the performance needs to be enhanced further.

More recently, following the remarkable progress in NLP [25], vision transformer (ViT) [26] first demonstrates that solely relying on self-attention mechanisms can surpass state-of-the-arts in image classification tasks when pre-trained on large-scale training data. Built upon the superior capability of self-attention in explicitly modeling long-range contextual interaction and learning abstract semantic information [27], many studies have been dedicated to introducing transformers into object segmentation tasks. SETR [28] is the first work that directly considers ViT as an encoder module in a semantic segmentation model. Strudel et al. build Segmenter [29] based on ViT encoder and mask transformer decoder to prove the potential of using pure transformers in semantic segmentation. Nevertheless, in practice, ViT is natively ineligible for pixel-wise dense prediction tasks, owing to its columnar outputs with coarse-grained resolution. [30][31] break this limitation by referring to the pyramid structure in CNNs, allowing a hierarchical feature representation for transformers under a limited computational burden. Fig.~\ref{fig1} empirically exhibits that pure transformer-based model, SETR [28], performs well on making category classification, but weaks in keeping fine spatial details. Additionally, these convolution-free transformers treat an image as a sequence of visual patches (tokens), lacking the inductive bias [32] of images in modeling local structures, such as edges and corners. Consequently, large-scale datasets and longer training periods are required to learn such inductive bias implicitly, going against the industrial imaging applications with limited samples. (see Section \uppercase\expandafter{\romannumeral5}).

Based on the above observation, it is evident that modeling long-distance dependency and capturing local structure information are both central issues in image segmentation. Hence, the latest efforts are being made to incorporate the merits of CNNs (local texture, inductive bias) and transformers (global context) into a unified framework, while there are still two challenges. First of all, the computational cost of self-attention is quadratic to the resolution of input image, which is prohibitive for pixel-wise prediction. Prior works [33][34] suggest calculating self-attention on lower-resolution CNNs feature map rather than the input images. Nevertheless, as the global interaction learning within low-level features is neglected, such simple combination will lead to suboptimal representation learning. After that, methods [35-37] make use of a pooling or convolutional operation to downsample the spatial scale of key and value in self-attention at the expense of some performance loss. Secondly, the self-attention is permutation-invariant [38], and fails in learning the order of the tokens in an input sequence. Previous studies [26][31] try to solve it by adding absolute and relative positional encoding, but will destroy the flexibility of the transformers. Recent advances in [39][40] encode the positional information with the padding of convolution operation only in one module of transformers, which tends to be not adequately effective yet. Furthermore, the above methods have made favorable progresses in general vision tasks, but exploring a specialized transformer-based model for pixel-wise surface defect detection is rarely studied.

Through the above analysis, we propose an efficient hybrid transformer architecture, called \textbf{Def}ect \textbf{T}ransformer (DefT), for surface defect detection in this paper, and its overall architecture follows the commonly-used encoder-decoder structure. To be specific, we firstly propose a convolutional stem block to effectively capture local spatial information. Next, to extract multi-scale features important for pixel-level defect segmentation, four patch aggregation blocks are successively adopted, each of which is followed by multiple DefT blocks responsible for learning locality and global dependencies simultaneously. Finally, a simple yet effective decoder module is utilized to recover detailed structures of the defects from the skip connections in the encoder. To the best of our knowledge, the proposed DefT network is the first work that explores transformer architecture for accurate pixel-wise surface defect detection.

Overall, our main contributions are summarized as follows:

1) we inventively reconsider the task of surface defect detection as a sequence-to-sequence prediction problem, and verify that transformer is a promising alternative to the CNNs for the pixel-wise defect detection tasks.

2) we propose an efficient hybrid transformer architecture by porting the properties of CNNs into the vision transformer, making the proposed DefT network inherit the advantage of transformers to model long-range dependencies and of CNNs to capture local structural relationships.

3) we conduct extensive experiments on three surface defect detection benchmarks to demonstrate the superior performance and generalization of our method, in comparison with current transformer- and CNN-based models.

The remainder of this paper is listed below. The related work is reviewed briefly in section \uppercase\expandafter{\romannumeral2}, and a detailed description of the proposed DefT network in section \uppercase\expandafter{\romannumeral3}. Next, section \uppercase\expandafter{\romannumeral4} presents a thorough comparison with the existing methods, and section \uppercase\expandafter{\romannumeral5} analyzes the contribution of key components of our model. Finally, the conclusion is drawn in Section \uppercase\expandafter{\romannumeral6}.

\section{Related works}
In this section, we briefly review two classes of defect detection methods, including traditional and CNN-based ones. Furthermore, the core network vision transformer is introduced in the last part.

\subsection{Traditional defect detection methods}
Traditional defect detection methods defined in this paper refer to the approaches that use hand-crafted features and prior knowledge. They can be mainly categorized into statistical, spectral, and model-based methods.
Statistical methods distinguish defects by evaluating the periodic distribution of statistical behaviors in the images, and the common statistics contain thresholding [2], gray-level [3], morphology [4], edge [5][6], et al. These methods are based on the assumption that the employed statistics are always stationary in the defect-free regions. Although the above statistical methods are the most popular among the early surface defect detection, they are likely to fail in the images with subtle intensity variation, such as noises and pseudo-defects.

Spectral methods argue that defects can be more easily recognized in a specific transform domain than the raw pixel domain. Therefore, multiple spatial-frequency domain features are utilized to better detect defects, such as wavelet transform [7], Fourier transform [8][9], Gabor transform [10], and so on. However, the performance of these approaches is heavily dependent on whether the transform domain selected is appropriate. Moreover, they have difficulty in representing textured surfaces with stochastic variations.

Model-based methods are able to identify defects from the images with complex textures by formulating image features distribution as a suitable mathematic model. Currently, there are mainly three models for defect detection, including markov random field model [11][12], active contour model [13] and RPCA model [14][15]. However, these methods suffer from considerable computational complexity. Meanwhile, setting feature descriptors manually tends to lack sufficient representation ability for different textures.

In brief, the above traditional methods are skilled at detecting defects from the background with simple texture, but their performance strongly depends on the selection of hand-crafted features and assumed priors, thus lacking generalization to handle different industrial scenarios.

\subsection{CNN-based defect detection methods}
Different from manual feature selection in traditional machine learning, recently emerging CNNs can automatically learn feature representation from low to high-level by stacking a series of convolutional layers. The low-level feature is beneficial to refine object details, and the high-level feature has consistent abstract semantics, which is crucially important for object detection tasks from the cluttered background. Accordingly, CNN-based methods, especially Fully Convolutional Neural Networks (FCNs) [41], have obtained significant advances in pixel-wise defect detection. Tong et al. [42] proposed an FCN-based model to enable end-to-end pixel-wise defect detection in pavement images. After that, UNet network [16] based on FCN encoder and stepwise decoder becomes dominated in object segmentation due to finer detail recovery. In [43][44], various UNet-style architectures were adopted for pavement crack detection and achieved state-of-the-arts compared with concurrent methods. However, the empirical receptive fields in the above methods are restricted by the size of the convolutional kernel, which prevents the accurate defects segmentation in challenging cases.

To capture more global context information, a lot of researches had been developed to enlarge the receptive field. In [19][45], convolution with large kernel was adopted to perceive contextual relation within a greater image region, which will bring considerable computational cost. Methods [21][46] introduced atrous convolution with different dilated rates to capture global context-aware information without increasing the number of parameters. Liu et al. [47] proposed an approach with free-form deformable receptive fields to more flexibly fit the shapes of the defects in metal surfaces. In addition, various attention modules were designed in [48][49] to adaptively reallocation feature responses by explicitly modeling interdependencies between the channels or spatial dimensionality. Nevertheless, plenty of empirical designs in these models are complicated and computationally demanding.

In summary, CNN-based methods have led to incredible progress in surface defect detection, but because of the inherent locality of convolution operations, they still gather information from local pixels in essence and lack the capability to model long-range contextual dependency explicitly.

\begin{figure*}[t]
  \centering
  \includegraphics[width=0.8\textwidth]{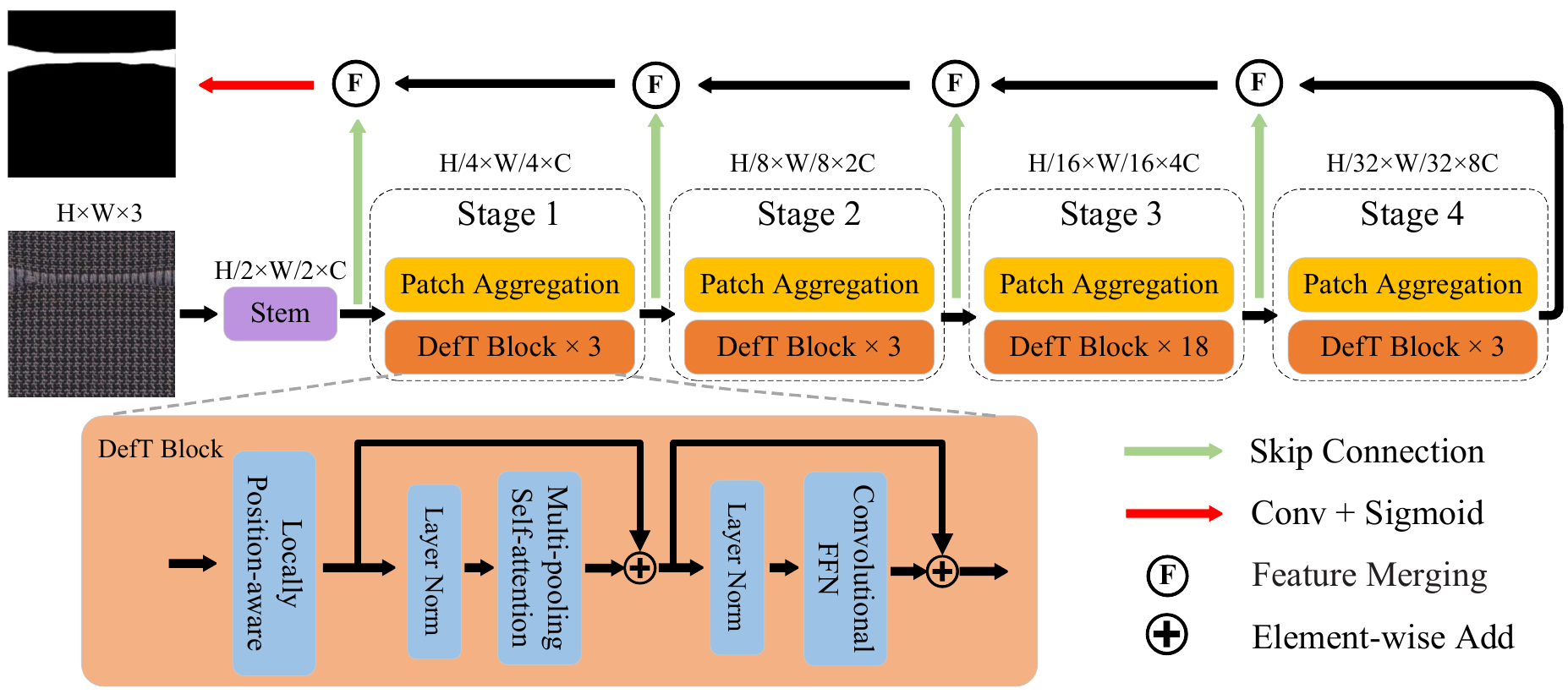}\\
  \caption{The overall architecture diagram of our proposed DefT network, which consists of two main modules, the lower one is the encoder module responsible for gradually reducing the spatial resolution and learning feature transformation, while the upper one is the decoder module used to progressively recover finer details by merging features from the corresponding encoder module.}\label{fig2}
\end{figure*}

\subsection{Transformer-based object detection methods}
Benefiting from the ability of self-attention mechanism within the transformer [25] to correlate every element with each other, thus allowing access to global interaction in the input sequence, there has been a broad interest to introduce transformers to computer vision in the past year. Dosovitskiy et al. [26] pioneeringly viewed an image as a sequence of patches (tokens) and fed them to a transformer encoder for image classification along with the position embedding. It delivered competitive performance compared to CNNs when training data is sufficient. Carion et al. [50] presented a novel end-to-end framework DETR based on transformers to locate objects without the need for NMS. For object segmentation, Zheng et al. [28] proposed the first study SETR that directly adopted ViT as an encoder module and combined with a simple decode. In [29], Segmenter is built based on ViT encoder and mask transformer decoder to demonstrate the feasibility of using pure transformers in semantic segmentation. However, ViT designed for image classification can only generate columnar low-resolution feature maps, resulting in poor segmentation performance. To adapt to pixel-wise prediction tasks, pyramid structure inspired by CNNs was introduced into the design of transformers to generate multi-level pyramid features [30][51].

However, there are still some performance gaps between the transformer- and CNN-based methods. One main reason is that convolution-free transformers neglect to preserve the locality crucial for fine spatial structures. Besides, pure transformers lack an intrinsic inductive bias [32] in modeling local visual structures, making the network difficult to converge. Hence, some recent methods suggested combining the transformers and CNNs into a unified architecture. Early studies [33][34] directly applied transformers upon the feature maps extracted from CNNs, but suffered from suboptimal representation learning. In addition, there are a lot of works trying to insert convolutional or pooling operations into the different modules of transformers. More specifically, CPVT [38] and ResT [39] presented a more flexible conditional positional encoding implemented by convolution with padding to replace the relative or absolute positional encodings in original transformers. In LocalViT [40] and CMT [52], a convolutional layer was inserted into the feed-forward network to leak spatial local information. Recent studies in [35][36][53] leveraged pooling or convolutional operation to reduce the spatial size of key and value in the self-attention calculation, which is also helpful for relieving high computational complexity. Xiao et al. [54] added convolutional operations on the top of the transformers to maintain the local continuity around the image patches.

For defect detection tasks using transformers, Dang et al. [55] proposed a DefectTR network based on DETR [50] for sewer defect localization and defect severity analysis. [56][57] introduced transformers into autoencoder to better reconstruct the deep features of images for anomaly detection in a semi-supervised way. As far as we know, few works adopt a fully supervised method based on transformers for high-quality pixel-wise surface defect detection.

\section{Methodology}
In this section, an overview of the proposed Defect Transformer (DefT) for surface defect detection is provided firstly. After that, the details of basic components are elaborately described successively. Finally, we present some training details of our network.

\subsection{Network Architecture}
Inspired by the success of CNNs in various dense prediction tasks and ViT in modeling long-range dependency, we subtly port the desirable properties of them into a unified architecture and explore an efficient hybrid transformer architecture for surface defect detection, the overall architecture is illustrated in Fig.~\ref{fig2}. Generally, the structure of our DefT network follows the prevalent UNet-like encoder-decoder networks. The encoder module shares a similar pipeline of ResNet [58], a convolutional stem block is applied firstly to retain more local information, followed by four stages to flexibly adjust the scale and dimension of the feature map, thus producing a hierarchical representation. Each stage consists of a patch aggregation block to achieve spatial downsampling, and a stack of DefT blocks, including a locally position-aware block (LPB), a lightweight multi-pooling self-attention (LMPS), and a convolutional feed-forward network (CFFN), to capture local and long-range information. Consequently, we can derive multi-scale pyramid features $\{F1, F2, F3, F4, F5\}$, whose strides are 2, 4, 8, 16, and 32 relative to the input, as shown in Table I. Afterwards, those hierarchical features are fed to the decoder module with skip connection to predict the segmentation mask.

\begin{table}[]
\centering
\setlength{\tabcolsep}{1.9mm}
\caption{Parameters setting of encoder module in our network architecture. The parameter of convolutional layers is represented as (kernel size$\times$kernel size)-(number of filters)-(stride). $H$ and $P$ denote the head number and various average pooling ratios in the LMPS respectively. E means the expansion ratio of the CFFN.}\label{table1}
\begin{tabular}{c|cc|c|c}
\hline
                         & \multicolumn{2}{c|}{Layer}                                                                        & Configuration                                                                              & Output Size            \\ \hline
Input                    & \multicolumn{2}{c|}{---}                                                                            & ---                                                                                          & $W{\rm{ \times }}H{\rm{ \times }}3$                  \\ \hline

Stem                     & \multicolumn{2}{c|}{\begin{tabular}[c]{@{}c@{}}Conv 1\\ Conv 2\\ Conv 3\end{tabular}}             & \begin{tabular}[c]{@{}c@{}}3$\times$3-C-2\\ 3$\times$3-C-1\\ 3$\times$3-C-1\end{tabular}                        & $\frac{W}{2}{\rm{ \times }}\frac{H}{2}{\rm{ \times}C}$                  \\ \hline

\multirow{4}{*}{Stage 1} & \multicolumn{2}{c|}{Patch Aggre}                                                                  & 3$\times$3-C-2                                                                                    & \multirow{4}{*}{$\frac{W}{4}{\rm{ \times }}\frac{H}{4}{\rm{ \times}C}$} \\ \cline{2-4}
                         & \multicolumn{1}{c|}{\multirow{3}{*}{\begin{tabular}[c]{@{}c@{}}DefT\\ Block\end{tabular}}} & LPB  & \multirow{3}{*}{\Bigg[ \begin{tabular}[c]{@{}c@{}}3$\times$3-C-1\\ H=1, E=4\\ P=\{12,16,20,24\} \end{tabular}\Bigg]$\times$3} &                        \\
                         & \multicolumn{1}{c|}{}                                                                      & LMPS &                                                                                            &                        \\
                         & \multicolumn{1}{c|}{}                                                                      & CFFN &                                                                                            &                        \\ \hline

\multirow{4}{*}{Stage 2} & \multicolumn{2}{c|}{Patch Aggre}                                                                  & 3$\times$3-2C-2                                                                                    & \multirow{4}{*}{$\frac{W}{8}{\rm{ \times }}\frac{H}{8}{\rm{ \times}2C}$} \\ \cline{2-4}
                         & \multicolumn{1}{c|}{\multirow{3}{*}{\begin{tabular}[c]{@{}c@{}}DefT\\ Block\end{tabular}}} & LPB  & \multirow{3}{*}{\Bigg[ \begin{tabular}[c]{@{}c@{}}3$\times$3-2C-1\\ H=2, E=4\\ P=\{6,8,10,12\} \end{tabular}\Bigg]$\times$3} &                        \\
                         & \multicolumn{1}{c|}{}                                                                      & LMPS &                                                                                            &                        \\
                         & \multicolumn{1}{c|}{}                                                                      & CFFN &                                                                                            &                        \\ \hline

\multirow{4}{*}{Stage 3} & \multicolumn{2}{c|}{Patch Aggre}                                                                  & 3$\times$3-4C-2                                                                                    & \multirow{4}{*}{$\frac{W}{16}{\rm{ \times }}\frac{H}{16}{\rm{ \times}4C}$} \\ \cline{2-4}
                         & \multicolumn{1}{c|}{\multirow{3}{*}{\begin{tabular}[c]{@{}c@{}}DefT\\ Block\end{tabular}}} & LPB  & \multirow{3}{*}{\Bigg[ \begin{tabular}[c]{@{}c@{}}3$\times$3-4C-1\\ H=4, E=4\\ P=\{3,4,5,6\} \end{tabular}\Bigg]$\times$18} &                        \\
                         & \multicolumn{1}{c|}{}                                                                      & LMPS &                                                                                            &                        \\
                         & \multicolumn{1}{c|}{}                                                                      & CFFN &                                                                                            &                        \\ \hline

\multirow{4}{*}{Stage 4} & \multicolumn{2}{c|}{Patch Aggre}                                                                  & 3$\times$3-8C-2                                                                                    & \multirow{4}{*}{$\frac{W}{32}{\rm{ \times }}\frac{H}{32}{\rm{ \times}8C}$} \\ \cline{2-4}
                         & \multicolumn{1}{c|}{\multirow{3}{*}{\begin{tabular}[c]{@{}c@{}}DefT\\ Block\end{tabular}}} & LPB  & \multirow{3}{*}{\Bigg[ \begin{tabular}[c]{@{}c@{}}3$\times$3-8C-1\\ H=8, E=4\\ P=\{1,2,3,4\} \end{tabular}\Bigg]$\times$3} &                        \\
                         & \multicolumn{1}{c|}{}                                                                      & LMPS &                                                                                            &                        \\
                         & \multicolumn{1}{c|}{}                                                                      & CFFN &                                                                                            &                        \\ \hline
\end{tabular}
\end{table}

\subsection{Encoder module}
\subsubsection{Stem Block}
In opposition to original patchify stem which directly splits the input image into patches of size 16$\times$16 [26] or 4$\times$4 [30,31], we perform smaller patches of size 2$\times$2 to retain finer spatial information, which is in favor of the pixel-wise surface defect detection. Moreover, as listed in Table~\ref{table1}, this process is implemented by the convolutional stem block, i.e. a 3$\times$3 convolution with stride 2 to halve the input resolution, followed by another two 3$\times$3 convolutions with stride 1 to facilitate network optimization and obtain better performance [54].

\subsubsection{Patch Aggregation Block}
Contrary to previous ViT that only generates single-scale coarse feature maps, the linear patch embedding layers are seminally introduced in PVT [30] to achieve spatial downsampling and representation enhancement as the network goes deeper. However, this process was initially accomplished by flattening, concatenating and linear projection on the non-overlapping image patches [30,31], making it prone to lose the spatial relation around neighboring patches. Instead, we use convolution to implement overlapping patch aggregation inspired by CvT [35]. Specifically, given an intermediate input of size $h \times w \times c$, we send it to go through a convolution with the kernel size $k$, stride $s$, padding size $p$ and kernel number $2c$. Accordingly, the size of shrunken output is $h' \times w' \times 2c$, where
\begin{equation}
\begin{array}{*{20}c}
   {h' = \frac{{h - k + 2p}}{s} + 1,} & {w' = \frac{{w - k + 2p}}{s} + 1}  \\
\end{array}
\end{equation}

In this paper, we set $k=3$, $s=2$, and $p=1$ respectively to perform patch aggregation block to produce 2x downsampling of resolution.

\subsubsection{DefT Block}
Defect transformer models local and long-range relations among the multi-scale features by using $L$ DefT blocks, each of which is mainly composed of a locally position-aware block (LPB), a lightweight multi-pooling self-attention (LMPS), and a convolutional feed-forward network (CFFN). Meanwhile, like the traditional transformer [25], layer normalization (LN) and residual connection are applied at the two ends of LMPS and CFFN respectively, as illustrated in Fig.~\ref{fig2}. Formally, the whole computation is written as

\begin{equation}
\begin{array}{l}
 {\rm{X}}_l^{{\rm{''}}} {\rm{=LPB(X}}_{l - 1} {\rm{)}} \\
 {\rm{X}}_l^{\rm{'}} {\rm{ = LMPS(LN(X}}_l^{{\rm{''}}} {\rm{)) + X}}_l^{{\rm{''}}}  \\
 {\rm{X}}_l {\rm{ = CFFN(LN(X}}_l^{\rm{'}} {\rm{)) + X}}_l^{\rm{'}}  \\
 \end{array}
\end{equation}
where ${\rm{X}}_l^{{\rm{''}}}$, ${\rm{X}}_l^{\rm{'}}$ and ${\rm{X}}_l$ represent the output of the LPB, LMPS and CFFN respectively for block $l$. Next, the following will introduce the details of these three parts.

\begin{figure}
  \centering
  \includegraphics[width=0.45\textwidth]{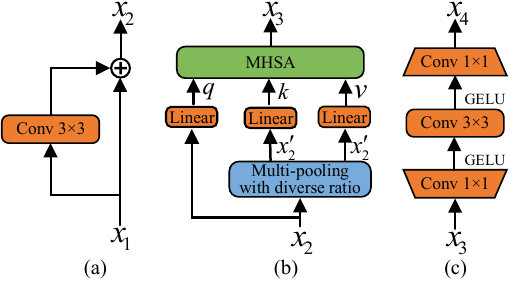}\\
  \caption{The structure of main components in the DefT block: (a) locally position-aware block; (b) lightweight multi-pooling self-attention; (c) convolutional feed-forward network.}\label{fig3}
\end{figure}

\paragraph{Locally position-aware block}
Unlike CNNs, the self-attention operation in transformer is inherently unable to perceive any positional information between the patches, which is essential to vision tasks. To ease this issue, many efforts [26][31] have been made to explicitly add absolute or relative positional encodings into each image patch. However, those methods are suboptimal and intricate [38], meanwhile, ignore spatial structural information. Considering that CNNs can implicitly encode position information from the zero-padding operation [59], we consider locally position-aware block to dynamically learn the local spatial information conditioned on the adjacent image patches. To be specific, as illustrated in Fig.~\ref{fig3}(a), let $x_1$ be the input, then the output $x_2$ is represent as
\begin{equation}
x_2  = x_1  + conv_{3 \times 3} (x_1 )
\end{equation}
where $conv_{3 \times 3} ( \cdot )$ means the learned position encoding derived from a 3$\times$3 convolution with zero padding 1. In this manner, we can implicitly embed the local spatial information into the input.

\paragraph{Lightweight multi-pooling self-attention}
Despite the strong capability of self-attention in capturing long-distance interaction, such original full self-attention, whose computational complexity is quadratic to the resolution of input images, is infeasible when handling high-resolution dense prediction tasks. Besides, the local spatial context would be ignored.

To address the above two issues, we apply the idea of pyramid pooling in CNNs onto the key and value of self-attention, forming the lightweight multi-pooling self-attention. Concretely, as depicted in Fig.~\ref{fig3}(b), given the input feature $x_2$, we first use the multiple average pooling operations with various ratios on it to produce the pyramid features [53], then flatten and concatenate them to obtain substitute $x'_2$. Mathematically, the output is calculated by
\begin{equation}
x'_2  = cat\left( {img2seq\left[ {avgpool_i \left( {x_2 } \right)|i \in P} \right]} \right)
\end{equation}
where $cat$ refers to a concatenation operation along the channel dimension, and $img2seq$ means a reshape operation to flatten the 2D image to a tensor sequence. $avgpool_i \left(  \cdot  \right)$ indicates the average pooling with ratio $i$. $P$ denotes the pooling ratios, and the values for different stages are listed in Table~\ref{table1}.

After that, unlike the traditional self-attention [26], the query $q$, key $k$ and value $v$ in our LMPS are represented as
\begin{equation}
\left( {q,k,v} \right) = \left( {x_2 w_q ,x'_2 w_k ,x'w_v } \right)
\end{equation}
where $w_q$, $w_k$ and $w_v$ are learnable parameters of linear projection respectively. Since the high pooling ratio, the output in Equ. (4) $x'_2$ has a shorter sequence while maintaining the contextual abstraction of input $x_2$, which provides efficiency via reducing the spatial scale of $k$ and $v$. Consequently, the attention calculation in our LMPS $x_3$ is computed as below
\begin{equation}
x_3  \buildrel \Delta \over = Attention\left( {q,k,v} \right) = {\rm{Softmax}}\left( {\frac{{qk^T }}{{\sqrt d }}} \right)v
\end{equation}
where ${\rm{Softmax}}\left(  \cdot  \right)$ is the softmax function used along the row vector, $d$ indicates the channel dimension of $k$ for scale balance. Noting that we suppose the head number in multi-head self-attention (MHSA) is 1 in the above analysis for simplicity, their actual values for different stages in our model are tabulated in Table~\ref{table1}. By this means, our LMPS not only reduces the computational load and memory usage, but also performs additional modeling of multi-scale contextual information.

\paragraph{Convolutional feed-forward network}
For feature transformation and non-linearity, a feed-forward network (FFN), containing two linear layers with an activation function in between, is generally employed after the self-attention operation. To further learn the locality mechanism, we insert a 3$\times$3 convolution with zero padding 1 again into FFN [40], named as the convolutional feed-forward network (CFFN). More precisely, as illustrated in Fig.~\ref{fig3}(c), the proposed CFFN is expressed as
\begin{equation}
x_4  = conv_{1 \times 1} (conv_{3 \times 3} (conv_{1 \times 1} (x_3 )))
\end{equation}
where the activation function GELU after the first two convolutions is omitted for convenience. The two $conv_{1 \times 1}$ indicate two 1$\times$1 convolutions functionally equivalent to two linear layers, which play a role in expanding and shrinking the feature dimensions respectively. While the $conv_{3 \times 3}$ means a convolution with kernel size 3 and padding size 1 for information interaction within a local region.

\subsection{Decoder module}
In the decoder module, we adopt skip-layer feature merging operation shown in Fig. 4 to gradually promote detail recovery and defect location. Technically, in each scale, the feature map $x$ from the decoder is upsampled twice and concatenated with the feature map $y$ from the symmetrical encoder stage, followed by a 3$\times$3 convolutional layer and an activation function to halves the channel dimension. This process can be formulated as
\begin{equation}
X = \phi (conv_{3 \times {\rm{3}}} (cat(up(x),y)))
\end{equation}
where $\phi$ is the ReLU activation function, $conv_{3 \times 3}$ represents 3$\times$3 convolutional layer, $cat$ means the concatenation operation, and $up$ denotes bilinear interpolation. In this way, we can successively refine defect details stage by stage, and reduce the information loss caused by downsampling. Finally, the last refined feature map is fed to a one-channel convolution layer and a sigmoid function to generate detection result.

\begin{figure}
  \centering
  \includegraphics[width=0.44\textwidth]{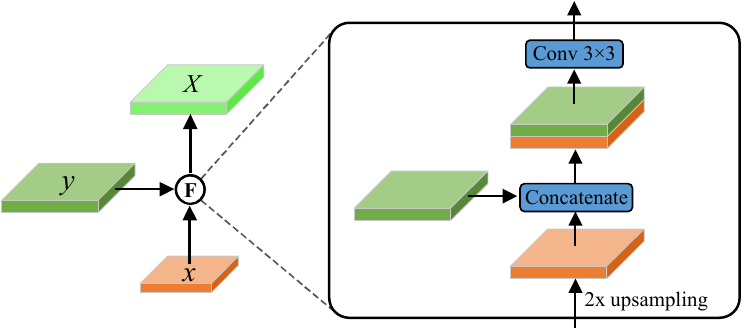}\\
  \caption{An illustration of skip-layer feature merging operation in the decoder module.}\label{fig4}
\end{figure}

Instead of using intricate components in [21][45], several simple alternations of upsampling operation and convolution layer utilized in this paper can yield impressive performance, which is attributed to powerful representations with local and global attention derived from the encoder module.

\subsection{Training strategy}

Our DefT network is an end-to-end pixel-wise defect detection network, and its specific process is shown in algorithm 1. The learnable parameters $\omega$ of the entire network are initialized by the default setting in PyTorch. During the training phase, all training samples are first resized to 256$\times$256 and cropped to 224$\times$224. Then, they are randomly fed into the network with batch size 8. We train the network using SGD optimizer with epoch 700, momentum 0.9, and weight decay 0.0001. The initial learning rate is empirically set to 0.003, and stepwise decreases following the $poly$ learning rate policy with the power of 0.9. Meanwhile, we adopt the hybrid loss in [21] for training, and deep supervision to speed convergence and boost performance. In inference, we uniformly scale the test image to 256$\times$256 and feed it into the network to predict detection result. All experiments in this paper are executed on a PC with an Intel i9-7900X processor and a NVIDIA P5000 GPU.
\begin{algorithm}[h]
\caption{~~~~~~Training of our DefT network}
\textbf{Input:} training$\_$dataset (img, label)\\
\textbf{Initialize:} DefT with default initialization, epoch, batch$\_$size, D = DataLoader(training$\_$dataset, batch$\_$size)\\
\textbf{for} ep \textbf{in range}(epoch)\\
\textcolor{white}{\ }\quad \textbf{for} (img, label) \textbf{in enumerate}(D) \\
    \textcolor{white}{\ }\qquad pred $\gets$ DefT(img$\left| \omega  \right.$)\\
    \textcolor{white}{\ }\qquad L $\gets$ Loss(pred, label)\\
    \textcolor{white}{\ }\qquad $\omega$ $\gets$ update(${{\partial L} \mathord{\left/ {\vphantom {{\partial L} {\partial w}}} \right. \kern-\nulldelimiterspace} {\partial w}}$)\\
\textcolor{white}{\ }\quad\textbf{end for}\\
\textbf{end for}\\
\textbf{Output:} DefT with optimal parameter DefT($\cdot \left| \omega  \right.$).
\end{algorithm}

\section{Experiments}
The experimental setup is first introduced in this section. Then, to demonstrate the superiority of our DefT, we report and analyze the comparison results with other deep learning models on three datasets.

\subsection{Experimental setups}
\subsubsection{Datasets Description}
~\\
\indent \textbf{SD-saliency-900 dataset} [21] consists of 900 steel surface defect images with high-quality annotation. Images in this dataset suffer from multiple difficulties, such as complex cluttered backgrounds, low contrast and noise interference. For comparison, we adopt the same data augmentation strategy as [21], where there are 810 and 900 images for training and testing, respectively.

\textbf{Fabric defect dataset} [60] provides the largest dataset for fabric defect detection with pixel-wise labels so far. Since the fabric wrinkle and weak defect features, this dataset becomes challenging. It comprises a total of 1600 images, which are split into 1200 and 400 for training and testing respectively.

\textbf{NRSD-MN dataset} [49] contains 3936 man-made images and 165 natural images collected from the no-service rail surface, most of which are challenging with diverse defect shapes and rust disturbance. To train various deep models, we randomly select 2971 man-made images for training, and the rest for testing.

\subsubsection{Datasets Description}
Along with the visual comparison, we follow the widely-used metrics to quantitatively evaluate detection performance, including False Positive Rate (FPR), False Negative Rate (FNR), Accuracy (ACC), Mean Absolute Error (MAE) and F1-measure (F1). FPR is a metric for meaning the ratio of background pixels falsely predicted as defects. FNR represents the ratio of defective pixels falsely identified as background. ACC indicates the proportion of the correctly predicted pixels against the total pixels. MAE is computed as the average absolute pixel-wise difference between the prediction result and the corresponding ground-truth map. F1 is an overall metric for performance evaluation. Obviously, the lower FPR, FNR, MAE, and the higher ACC, F1, the better performance. These metrics are defined as follows
$$\begin{array}{*{20}c}
   {{\rm{FPR = }}{{{\rm{FP}}} \mathord{\left/
 {\vphantom {{{\rm{FP}}} {\left( {{\rm{FP + TN}}} \right)}}} \right.
 \kern-\nulldelimiterspace} {\left( {{\rm{FP + TN}}} \right)}},} & {{\rm{FNR = }}{{{\rm{FN}}} \mathord{\left/
 {\vphantom {{{\rm{FN}}} {\left( {{\rm{FN + TP}}} \right)}}} \right.
 \kern-\nulldelimiterspace} {\left( {{\rm{FN + TP}}} \right)}}}  \\
\end{array}$$
$${\rm{ACC = }}{{\left( {{\rm{TP + TN}}} \right)} \mathord{\left/
 {\vphantom {{\left( {{\rm{TP + TN}}} \right)} {\left( {{\rm{TP + FP + TN + FN}}} \right)}}} \right.
 \kern-\nulldelimiterspace} {\left( {{\rm{TP + FP + TN + FN}}} \right)}}$$
$$F1 = {{2TP} \mathord{\left/
 {\vphantom {{2TP} {\left( {2TP + FP + FN} \right)}}} \right.
 \kern-\nulldelimiterspace} {\left( {2TP + FP + FN} \right)}}$$
$${\rm{MAE}} = \frac{1}{{{\rm{w}} \times {\rm{h}}}}\sum\nolimits_{{\rm{i}} = 1}^{\rm{w}} {\sum\nolimits_{{\rm{j}} = 1}^{\rm{h}} {\left| {{\rm{p}}\left( {{\rm{i}},{\rm{j}}} \right) - {\rm{g}}\left( {{\rm{i}},{\rm{j}}} \right)} \right|} }$$
where FP, TN, FN, TP denote the number of false positive, true negative, false negative and true positive pixels, respectively. $p$ is the predicted result and $g$ is its ground truth. $w$ and $h$ indicate the width and height of images, respectively. Additionally, we also report the total parameters, theoretical computational complexity (FLOPs) and inference speed for efficiency comparison. Furthermore, Precision-Recall (PR) and F-measure curves are provided to evaluate models more intuitively.

\subsubsection{Competing methods}
We compare the proposed DefT network with eight advanced deep learning methods: 1) pure transformer-based methods SETR [28] and Swin-unet [51]; 2) hybrid transformer-based methods TransUNet [33], SegFormer [36] and MedT [61]; 3) CNN-based generic object detector UNet [16]; as well as 4) CNN-based defect detectors MCnet [49] and EDRNet [21]. For a fair comparison, these methods are all retrained to adapt to different defect detection tasks.

\subsection{Experimental results}

\subsubsection{Comparison on SD-saliency-900 dataset}
\begin{figure}[h]
  \centering
  \includegraphics[width=0.50\textwidth]{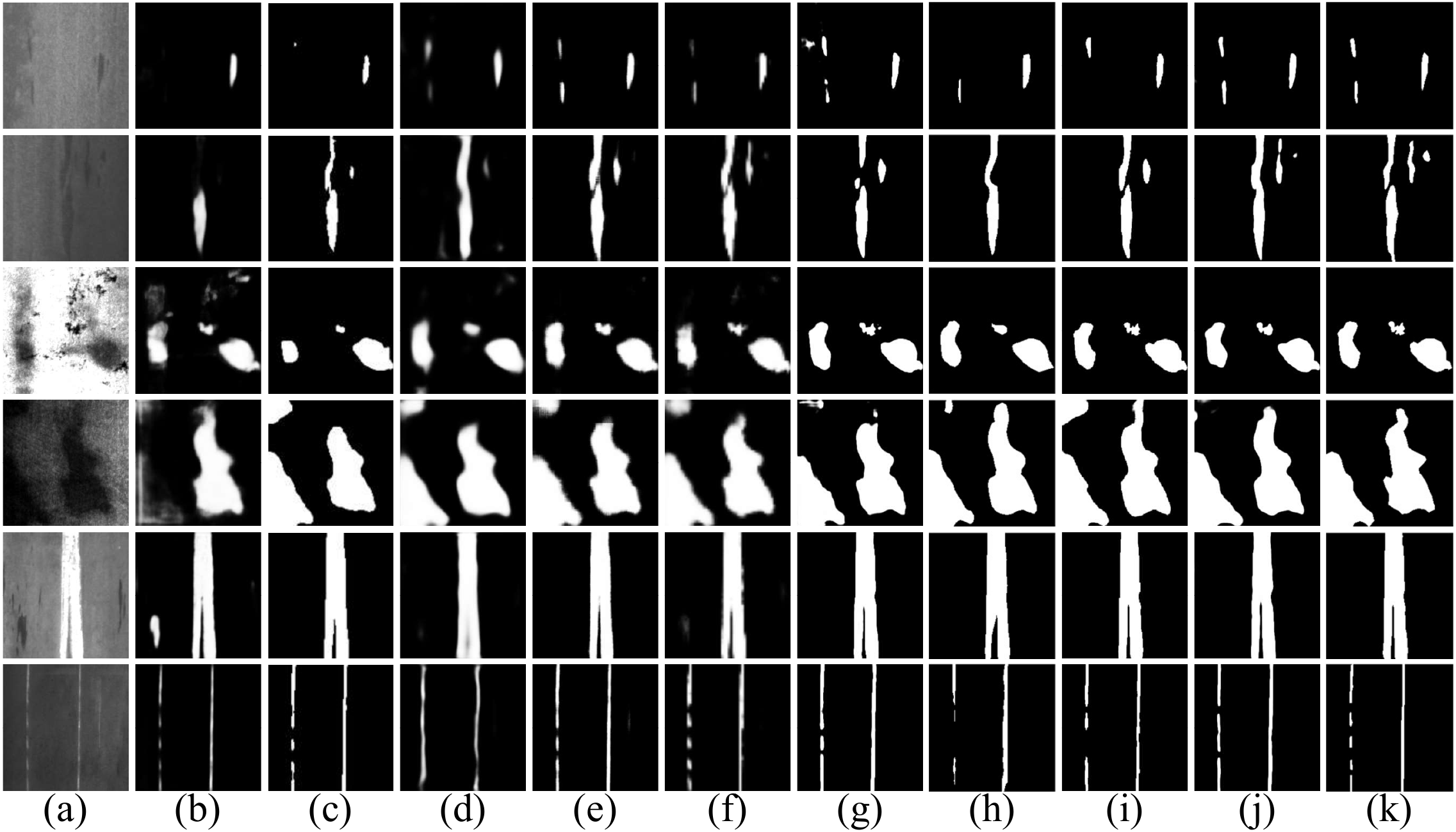}\\
  \caption{Visual comparison of different methods on SD-saliency-900 dataset. (a) Original image. (b) UNet. (c) MedT. (d) SETR. (e) Swin-unet. (f) SegFormer. (g) TransUNet. (h) MCnet. (i) EDRNet. (j) Ours. (k) Ground truth.}\label{fig5}
\end{figure}

The visual comparison of our method and other methods for partial SD-saliency-900 dataset is shown in Fig.~\ref{fig5}, and corresponding quantitative results are listed in Table~\ref{table2}. We can see that UNet [16] is likely to erroneously detect some impurities, meanwhile, miss some defects with low contrast, thus giving rise to a high FPR and the largest FNR, as shown in Table~\ref{table2}. While SETR [28] and Swin-unet [51] can perform better in recognizing the defects from the complex background, but produce results with blurry boundaries. As for the results of MedT [61], SegFormer [36] and TransUNet [33], they possess promising performance in various challenging cases, but there are still some gaps compared with specialized defect detectors MCnet [49] and EDRNet [21]. In addition, detection results of our method are much close to the corresponding ground-truth maps, they can not only highlight the defect regions precisely but also suppress the background clutters well. Beyond visual superiority, our method substantially outperforms all the competing methods in terms of FNR, F1 and ACC.

\begin{table}[h]
\centering
\setlength{\tabcolsep}{1.4mm}
\caption{Quantitative evaluation of different methods on SD-saliency-900 dataset. Values highlighted in red and blue are the top two performances.}\label{table2}
\begin{tabular}{clccccc}
\hline
Dataset                          & \multicolumn{1}{l}{Methods} & FPR    & FNR    & F1     & ACC   & MAE    \\ \hline
\multirow{9}{*}{SD-saliency-900} & UNet [16]               & .0186 & .3719 & .7093 & .9406 & .0415 \\
                                 & MedT [61]               & .0143 & .3336 & .7695 & .9284 & .0296 \\
                                 & SETR [28]               & .0304 & .2469 & .7567 & .9450 & .0397 \\
                                 & Swin-unet [51]          & .0209 & .1336 & .8641 & .9643 & .0242 \\
                                 & SegFormer [36]          & .0265 & .1779 & .8287 & .9525 & .0312 \\
                                 & TransUNet [33]          & .0165 & \textcolor{blue}{.0648} & .9090 & .9777 & .0165 \\
                                 & MCnet [49]              & \textcolor{red}{.0130} & .1875 & .8520 & .9660 & .0176 \\
                                 & EDRNet [21]             & .0133 & .0705 & \textcolor{blue}{.9107} & \textcolor{blue}{.9783} & \textcolor{red}{.0132} \\
                                 & Ours                        & \textcolor{blue}{.0131} & \textcolor{red}{.0641} & \textcolor{red}{.9162} & \textcolor{red}{.9791} & \textcolor{blue}{.0134} \\ \hline
\end{tabular}
\end{table}

\subsubsection{Comparison on Fabric defect dataset}
\begin{figure}[h]
  \centering
  \includegraphics[width=0.491\textwidth]{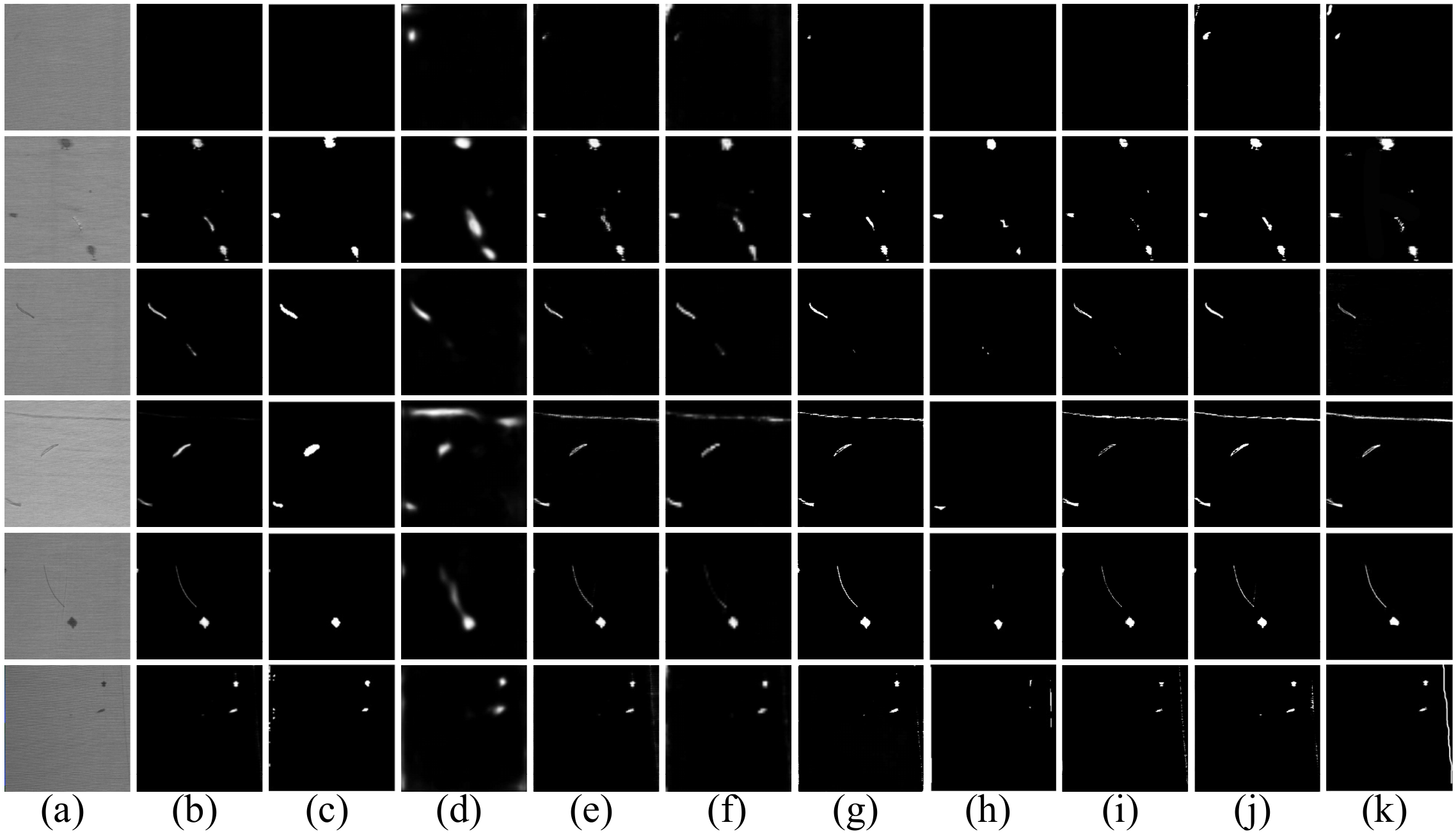}\\
  \caption{Visual comparison of different methods on fabric defect dataset. (a) Original image. (b) UNet. (c) MedT. (d) SETR. (e) Swin-unet. (f) SegFormer. (g) TransUNet. (h) MCnet. (i) EDRNet. (j) Ours. (k) Ground truth.}\label{fig6}
\end{figure}

Fig.~\ref{fig6} and Table~\ref{table3} illustrates the visual and quantitative comparison of different methods on fabric defect dataset respectively. As we can observed in the second column of Fig.~\ref{fig6}, UNet [16] is easy to miss some defects with weak features, and over-detect when dealing with fabric wrinkles. Although pure transformer-based methods SETR [28] and Swin-unet [51] have obvious advantages over UNet [16] in identifying error-prone wrinkles, they still tend to ignore some defect regions and do not have clear boundary. Existing hybrid transformer models MedT [61], SegFormer [36] and TransUNet [33] are not specifically designed for defect detection, they produce competitive results, but there is still room for improvement in small defects recognition. Relying on the complicated design in the decoder module, defect detection models MCnet [49] and EDRNet [21] can perform well in most cases at the cost of more computation. In contrast, just utilizing a simple decoder module with less computational cost, our DefT is superior or comparable to them. Numerically, from Table~\ref{table3}, our DefT achieves the best or second best performance on five evaluation metrics.

\begin{table}[h]
\centering
\setlength{\tabcolsep}{1.4mm}
\caption{Quantitative evaluation of different methods on fabric defect dataset. Values highlighted in red and blue are the top two performances.}\label{table3}
\begin{tabular}{clccccc}
\hline
Dataset                        & Methods            & FPR    & FNR    & F1     & ACC    & MAE    \\ \hline
\multirow{9}{*}{Fabric defect} & UNet {[}16{]}      & .0064 & .7113 & .3887 & .9715 & .0051 \\
                               & MedT {[}60{]}      & .0082 & .8319 & .2497 & .9555 & .0083 \\
                               & SETR {[}28{]}      & .0185 & .6352 & .3317 & .9680 & .0160 \\
                               & Swin-unet {[}51{]} & .0100 & .6174 & .4671 & .9673 & .0071 \\
                               & SegFormer {[}36{]} & .0104 & .6225 & .4596 & .9664 & .0081 \\
                               & TransUNet {[}33{]} & .0090 & \textcolor{blue}{.5503} & .5016 & .9771 & .0056 \\
                               & MCnet {[}49{]}     & \textcolor{red}{.0053} & .7137 & .2973 & \textcolor{blue}{.9775} & .0059 \\
                               & EDRNet {[}21{]}    & \textcolor{blue}{.0074} & .6202 & \textcolor{blue}{.4837} & .9699 & \textcolor{red}{.0047} \\
                               & Ours               & \textcolor{red}{.0053} & \textcolor{red}{.4874} & \textcolor{red}{.5448} & \textcolor{red}{.9808} & \textcolor{blue}{.0049} \\ \hline
\end{tabular}
\end{table}

\subsubsection{Comparison on NRSD-MN dataset}
\begin{figure}[h]
  \centering
  \includegraphics[width=0.5\textwidth]{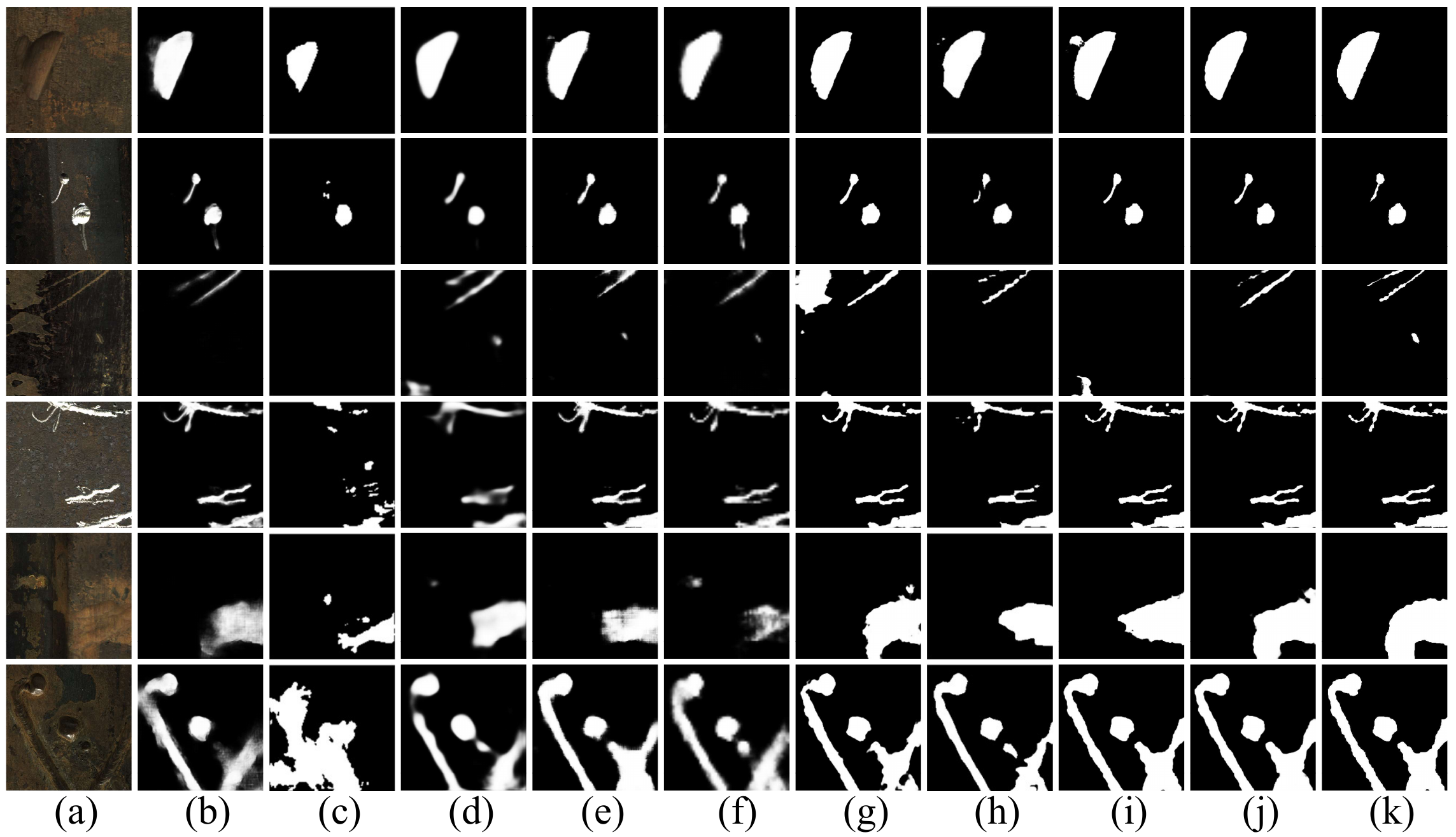}\\
  \caption{Visual comparison of different methods on NRSD-MN dataset. (a) Original image. (b) UNet. (c) MedT. (d) SETR. (e) Swin-unet. (f) SegFormer. (g) TransUNet. (h) MCnet. (i) EDRNet. (j) Ours. (k) Ground truth.}\label{fig7}
\end{figure}

Sample surface images on NRSD-MN dataset and corresponding prediction results are visualized in Fig.~\ref{fig7}, and the quantitative results of various models are tabulated in Table~\ref{table4}. It can be clearly observed that UNet [16] can detect most of the defects, but tends to regard some rusts as defect objects, and the detected defect areas are incomplete. In contrast, Swin-unet [28] and SETR [51] are less likely to false detection, but fail to generate clear structures in detail. We also observe that the methods combining CNN and transformer, e.g. MedT [61], SegFormer [36] and TransUNet [33], obtain consistently better local details and improve detection results, but they are still not on par with state-of-the-art MCNet [49] and EDRNet [21]. Compared to the above methods, our method can restrain the influence of the rusts and yield comparable accuracy to the state-of-the-arts only with a simple decoder module. It is mainly attributed to the excellent feature representation ability of our encoder module, particularly in learning global information interactions. In general, from Table~\ref{table4}, our method gets a decent improvement of 0.5$\%$ and 0.04$\%$ in F1 and ACC, while yields decline of 4.5$\%$ and 1.2$\%$ in FNR and MAE compared with the second best.

\begin{table}[h]
\centering
\setlength{\tabcolsep}{1.4mm}
\caption{ Quantitative evaluation of different methods on NRSD-MN dataset. Values highlighted in red and blue are the top two performances.}\label{table4}
\begin{tabular}{clccccc}
\hline
Dataset                  & Methods            & FPR    & FNR    & F1     & ACC    & MAE    \\ \hline
\multirow{9}{*}{NRSD-MN} & UNet {[}16{]}      & .0141 & .3269 & .7482 & .9545 & .0348 \\
                         & MedT {[}60{]}      & .0199 & .7781 & .2959 & .9296 & .0757 \\
                         & SETR {[}28{]}      & .0160 & .2799 & .7683 & .9592 & .0344 \\
                         & Swin-unet {[}51{]} & .0135 & .2248 & .8162 & .9658 & .0281 \\
                         & SegFormer {[}36{]} & .0163 & .2456 & .7967 & .9597 & .0318 \\
                         & TransUNet {[}33{]} & .0119 & \textcolor{blue}{.2045} & \textcolor{blue}{.8297} & \textcolor{blue}{.9711} & .0254 \\
                         & MCnet {[}49{]}     & \textcolor{red}{.0100} & .2621 & .8055 & .9659 & \textcolor{blue}{.0245} \\
                         & EDRNet {[}21{]}    & .0117 & .2206 & .8205 & .9705 & .0247 \\
                         & Ours               & \textcolor{blue}{.0115} & \textcolor{red}{.1952} & \textcolor{red}{.8337} & \textcolor{red}{.9715} & \textcolor{red}{.0242} \\ \hline
\end{tabular}
\end{table}

\subsubsection{Efficiency comparison}
This section compares the parameter size and inference time of different methods, the related results are listed in Table~\ref{table5}. Overall, our DefT achieves remarkable detection performance with less or close parameters and computations, thus having a better speed-accuracy trade-off. The total parameter size of the DefT is 30.56M, which is only larger than MedT and SegFormer, but our inference speed and segmentation accuracy consistently outperform both of them. Besides, our detection model DefT can handle a test image at a speed of 0.17s without accelerated implementations such as TensorRT, which is not the shortest but still acceptable in practice. The above results turn out that our method can better meet the needs of high efficiency and high accuracy in actual production.

\begin{table}[h]
\centering
\caption{Comparison of parameter size and inference speed of different methods.}\label{table5}
\begin{tabular}{c|ccccc}
\hline
Methods  & MedT  & TransUNet & SegFormer & Swin-Unet & SETR   \\
Params (M) & 1.56  & 105.28    & 27.35     & 48.194    & 105.83 \\
Times (s) & 0.37  & 0.18      & 0.17      & 0.17      & 0.16   \\ \hline
Methods  & UNet  & MCnet     & EDRNet    & Ours      &        \\
Params (M) & 34.53 & 42.30     & 39.31     & 30.56     &        \\
Times (s) & 0.09  & 0.10      &  0.13       & 0.17      &        \\ \hline
\end{tabular}
\end{table}

\section{Discussion}
An ablation study is carried out in this section to evaluate each key component of the proposed method. Beyond ablation, we also discuss the influence of several critical components in our model.

\subsection{Ablation Analysis}
To validate the contribution of each component in our DefT network, a series of ablation experiences are conducted on the SD-saliency-900 dataset. We take the model based on PVT [30]-like encoder and our decoder as the baseline, and integrate one component upon the baseline at a time, i.e., convolutional stem block (CSB), patch aggregation block (PAB), LPB, LMPS, and CFFN. All these variants are trained using identical training strategy.

\begin{table}[h]
\centering
\caption{Comparison of different configurations on the SD-saliency-900 dataset. FLOPs is calculated with the input size of 256$\times$256.}\label{table6}
\setlength{\tabcolsep}{1.1mm}
\begin{tabular}{l|c|c}
\hline
Configurations                        & MAE    & FLOPs (G) \\ \hline
Baseline                              & 0.0185 & 6.74      \\
Baseline+CSB                          & 0.0163 & 8.71      \\
Baseline+CSB+PAB                      & 0.0154 & 8.87      \\
Baseline+CSB+PAB+LPB                  & 0.0146 & 8.89      \\
Baseline+CSB+PAB+LPB+LMPS             & 0.0136 & 8.63      \\
Baseline+CSB+PAB+LPB+LMPS+CFFN (Ours) & 0.0134 & 8.72      \\ \hline
\end{tabular}
\end{table}

Given the results shown in Table~\ref{table6}, one can see that each component contributes to performance enhancement to a certain degree, our DefT network integrating all components gets the best detection results with an affordable computational overhead. More specifically, going from a patch size of 4 to 2 in our CSB, we observe a significant improvement of 11.9$\%$ in MAE at the cost of a little more computational complexity. Then, the linear patch embedding layer in PVT [30] is replaced with our PAB to preserve spatial relation, this simple alteration leads to an obvious performance gain (+ 5.5$\%$ in MAE) with a negligible extra computation. Next, the addition of LPB brings a relative performance boost of 5.2$\%$ in MAE with nearly the same computational cost. Furthermore, when combined with our LMPS, such variant further improves the performance (0.0146 v.s. 0.0136) in terms of MAE even with less computational overhead, suggesting the superiority and high efficiency of LMPS. Finally, the replacement of FFN in the baseline by our CFFN reaches a new state-of-the-art.

\begin{figure}[h]
  \centering
  \includegraphics[scale=0.293]{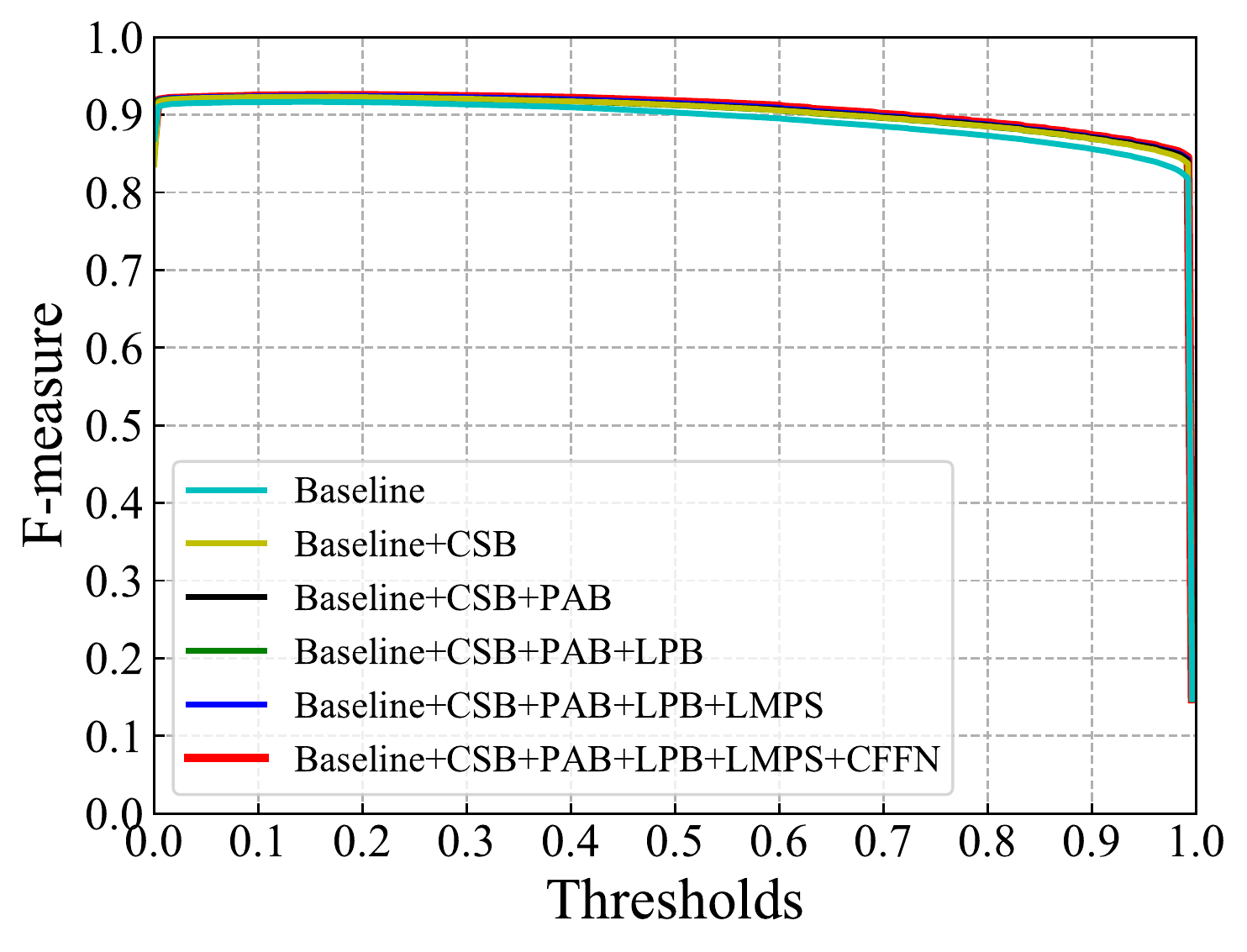}
  \hspace{0.001in}
  \includegraphics[scale=0.293]{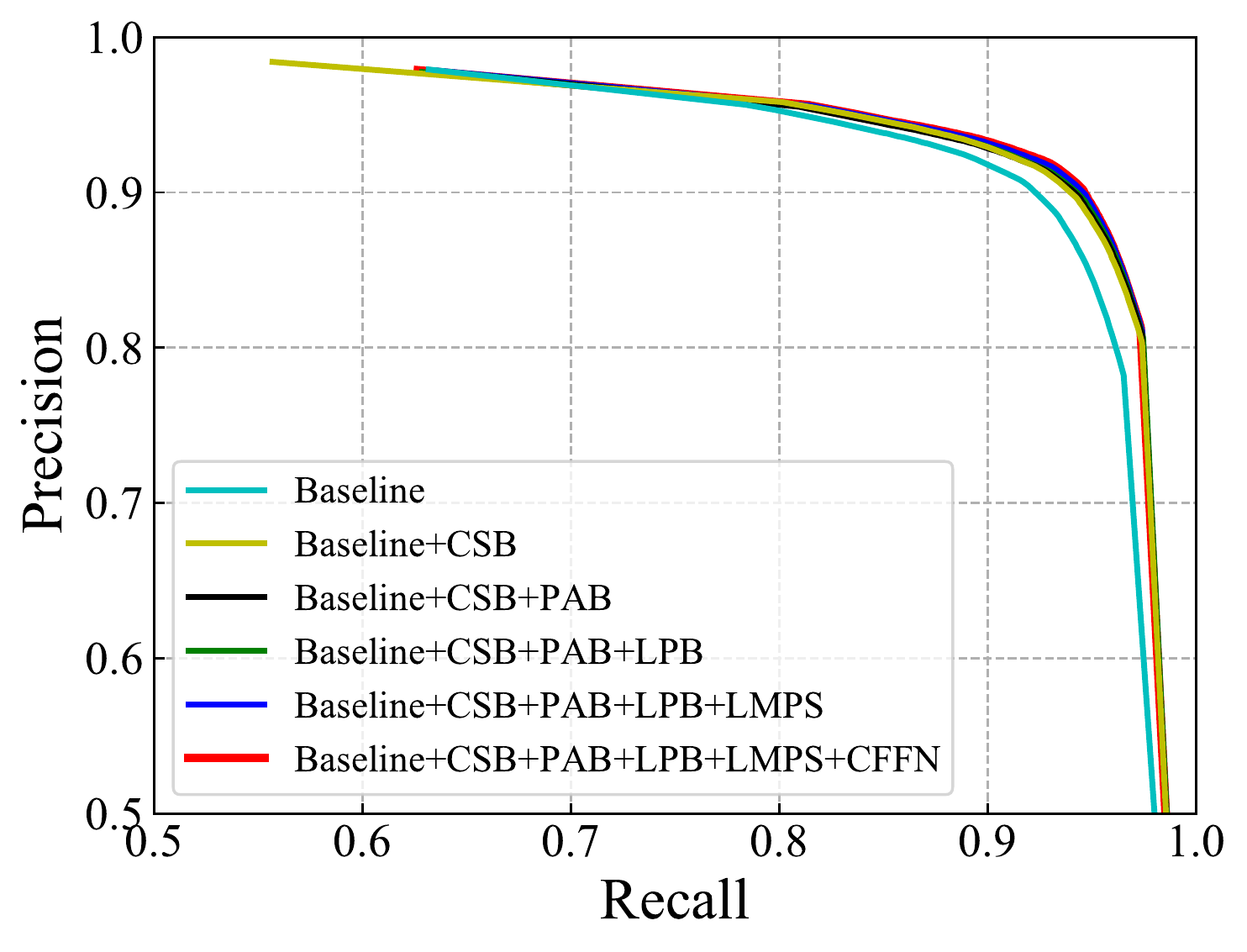}
  \caption{Performance evaluation of different configurations in terms of F-measure (left) and PR (right) curves.}\label{fig8}
\end{figure}

Meanwhile, the curves of PR and F-measure reported in Fig.~\ref{fig8} more intuitively demonstrate that all components in our DefT model are necessary to obtain the best detection performance.

\subsection{Effect of position encoding}
\begin{table}[h]
\centering
\caption{Our DefT with different position encoding methods.}\label{table7}
\begin{tabular}{l|cccc}
\hline
Type    & APE   & RPE   & Ours  & N/A   \\ \hline
ACC(\%) & 96.93 & 97.31 & 98.08 & 97.12 \\ \hline
\end{tabular}
\end{table}
In this section, we compare different types of position encoding methods on the fabric defect dataset: 1) absolute position encoding used in ViT [26] (APE); 2) relative position encoding used in Swin transformer [31] (RPE); 3) LPB and CFFN specially designed in our DefT (Ours). As tabulated in Table~\ref{table7}, our method surpasses the models with APE and RPE by 1.15$\%$ and 0.77$\%$ on detection accuracy, indicating that local position information can be better encoded by convolution operation with padding used in our method. Furthermore, it is exciting to find that our model without specialized position encoding N/A (i.e. disabling LPB and replacing CFFN with FFN) is still competitive. We believe that it is because the convolution ideas introduced in CSB, PAB and LMPS can also aid the model in making sense of local spatial context. This further confirms the importance of porting the properties of CNN into transformer for surface defect detection.

\subsection{Analysis of model efficiency}
There are two merits of CNNs that our network inherited, one is finer local details that had been discussed, and the other is inductive bias helpful for training and data efficiency. To better understand the impact of inductive bias, we set two training settings to compare our hybrid transformer DefT with pure transformer Swin-Unet: i) training with the full NRSD-MN dataset for the {60, 300, 510, 700} epoch; and ii) training with the {30$\%$, 60$\%$, 100$\%$} NRSD-MN dataset for the full epoch.

\begin{figure}[h]
  \centering
  \includegraphics[width=0.48\textwidth]{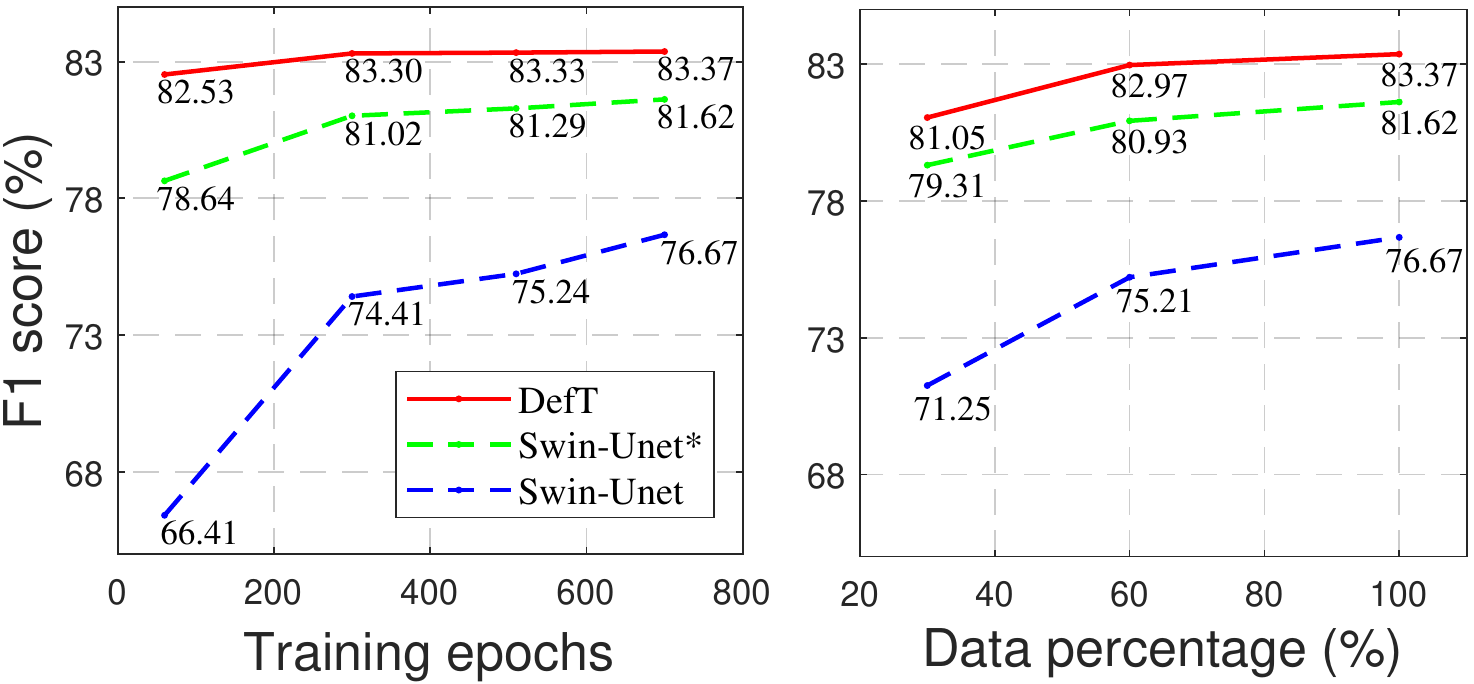}\\
  \caption{Comparison of training efficiency (left) and data efficiency (right) of DefT and Swin-Unet on NRSD-MN dataset. * means the model initialized by pre-trained weights on ImageNet.}\label{fig9}
\end{figure}

As can be observed in Fig.~\ref{fig9}, our DefT network consistently surpasses Swin-Unet by a large margin regarding training efficiency (left) and data efficiency (right). The training efficiency of our model is embodied by its less sensitivity of performance to the number of training epochs. For example, Swin-Unet yields a relative decline of 13.38$\%$ in F1 score when reducing the training epoch from 700 to 60, while our model is only 1.01$\%$. Besides, our DefT trained for only 60 epochs achieve an 82.53$\%$ F1, which significantly surpasses Swin-Unet trained for the full epochs, and even outperforms its pre-trained counterpart Swin-Unet*. The favorable training efficiency of our DefT is reflected in its extremely rapid convergence. Due to the intrinsic inductive bias in our method, DefT shows high stability compared with Swin-Unet when reducing the amount of training data. Alternatively, Swin-Unet* learns the inductive bias from amounts of data (ImageNet) implicitly, so it also shows comparable stability. It is also worth noting that our model only with 30$\%$ training data can obtain competitive performances compared to Swin-Unet* with 100$\%$ data. We speculate the reason is that explicit learning inductive bias from convolution is much more effective than learning from large-scale dataset implicit.

\section{Conclusion}
In this paper, we present the first study to explore the application of transformers for accurate pixel-wise surface defect detection. To this end, we combine transformer with CNN to construct a powerful encoder module, which enjoys the benefits of high-level global contextual information and low-level local details simultaneously. Then, the encoder module is coupled with a simple yet effective decoder via skip connections, thus forming an efficient hybrid transformer architecture for defect detection named DefT. Experimental results over the three defect datasets show that the proposed method is comparable to or outperforms existing transformer- and CNN-based methods in terms of detection performance, data efficiency, training efficiency, and generalization.





\begin{IEEEbiography}[{\includegraphics[width=1in,height=1.25in,clip,keepaspectratio]{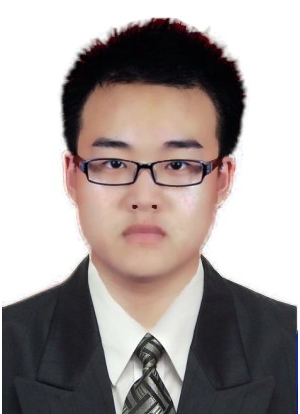}}]{Junpu Wang} received the B.S. and M.S. degree in signal and information processing from the School of Electronic and Information Engineering, from the Zhongyuan University of Technology, Zhengzhou, China, in 2016 and 2019, respectively. He is currently pursuing the Ph.D. degree with School of Automation Engineering, Nanjing University of Aeronautics and Astronautics. His research interests include image processing, and surface defect detection.
\end{IEEEbiography}

\begin{IEEEbiography}[{\includegraphics[width=1in,height=1.25in,clip,keepaspectratio]{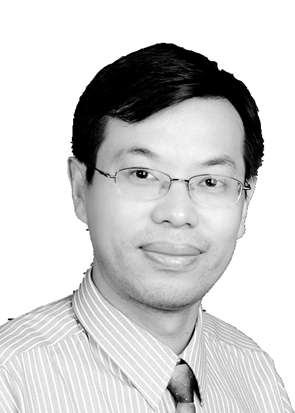}}]{Guili Xu} received the Ph.D. degree from Jiangsu University, in 2002. He is currently a Professor with the College of Automation Engineering, Nanjing University of Aeronautics and Astronautics (NUAA), Nanjing, China. He has authored over 100 publications. His current research interests are photoelectric measurement, computer vision, and intelligent systems.
\end{IEEEbiography}

\begin{IEEEbiography}[{\includegraphics[width=1in,height=1.25in,clip,keepaspectratio]{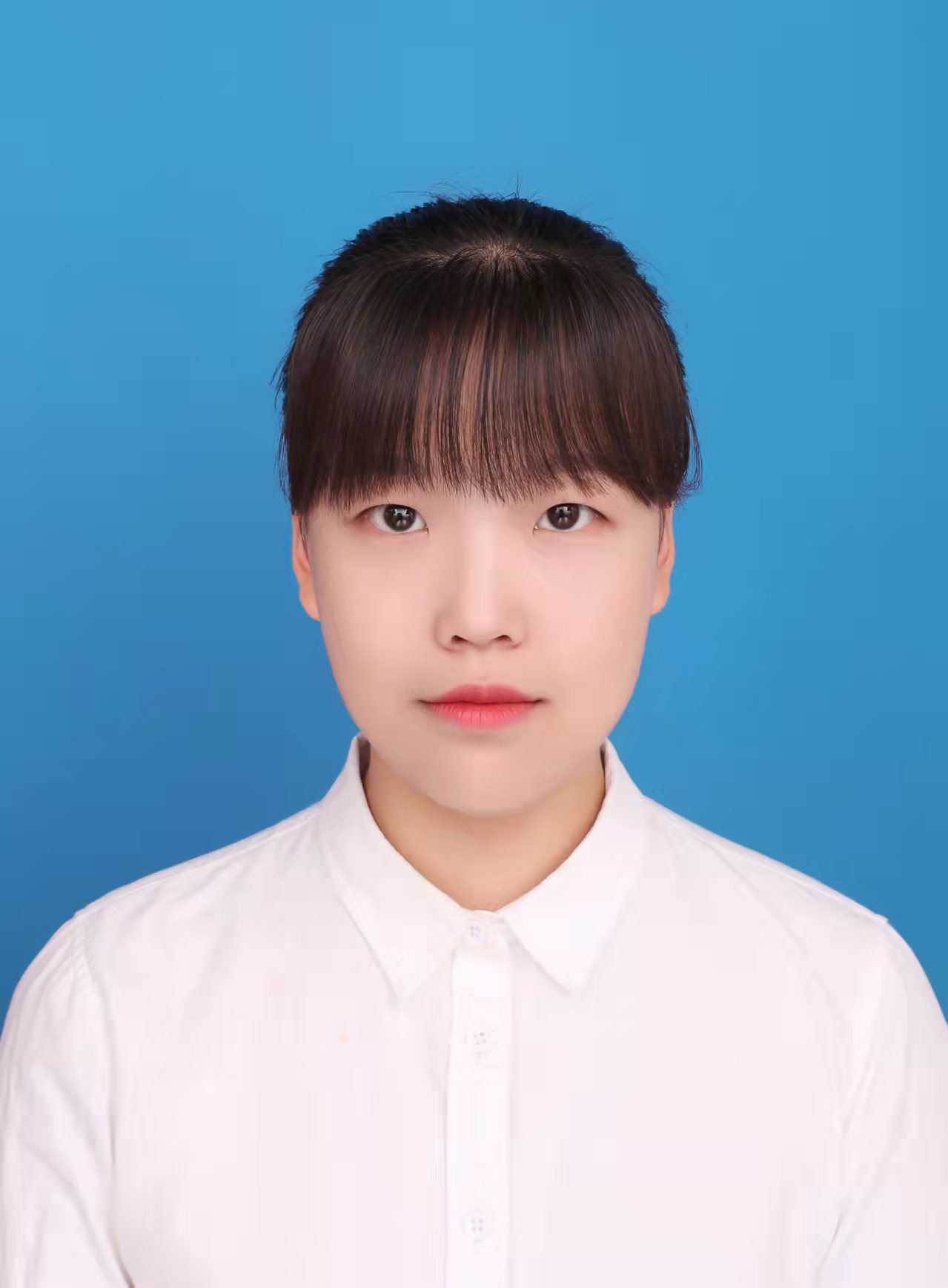}}]{Jinjin Wang} received her master's degree from Zhongyuan University of Technology in Zhenzhou,China in 2020. At present, she is studying for the doctorate at Nanjing University of Aeronautics and Astronautics. Her research interests include deep learning, hyperspectral classification, and image processing.

\end{IEEEbiography}

\begin{IEEEbiography}[{\includegraphics[width=1in,height=1.25in,clip,keepaspectratio]{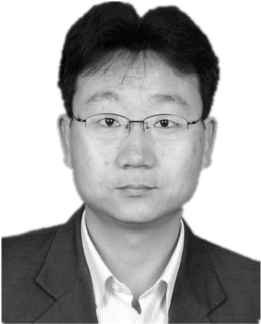}}]{Zhengsheng Wang} received the Ph.D. degree in computational mathematics from the Nanjing University of Aeronautics and Astronautics (NUAA), Nanjing, China, in 2006. He is currently a Professor and the Associate Dean of the College of Science, NUAA. His research interests focus on numerical algebra and applications, matrix methods in data mining and pattern recognition, and computational methods in science and engineering.
\end{IEEEbiography}

\begin{IEEEbiography}[{\includegraphics[width=1in,height=1.25in,clip,keepaspectratio]{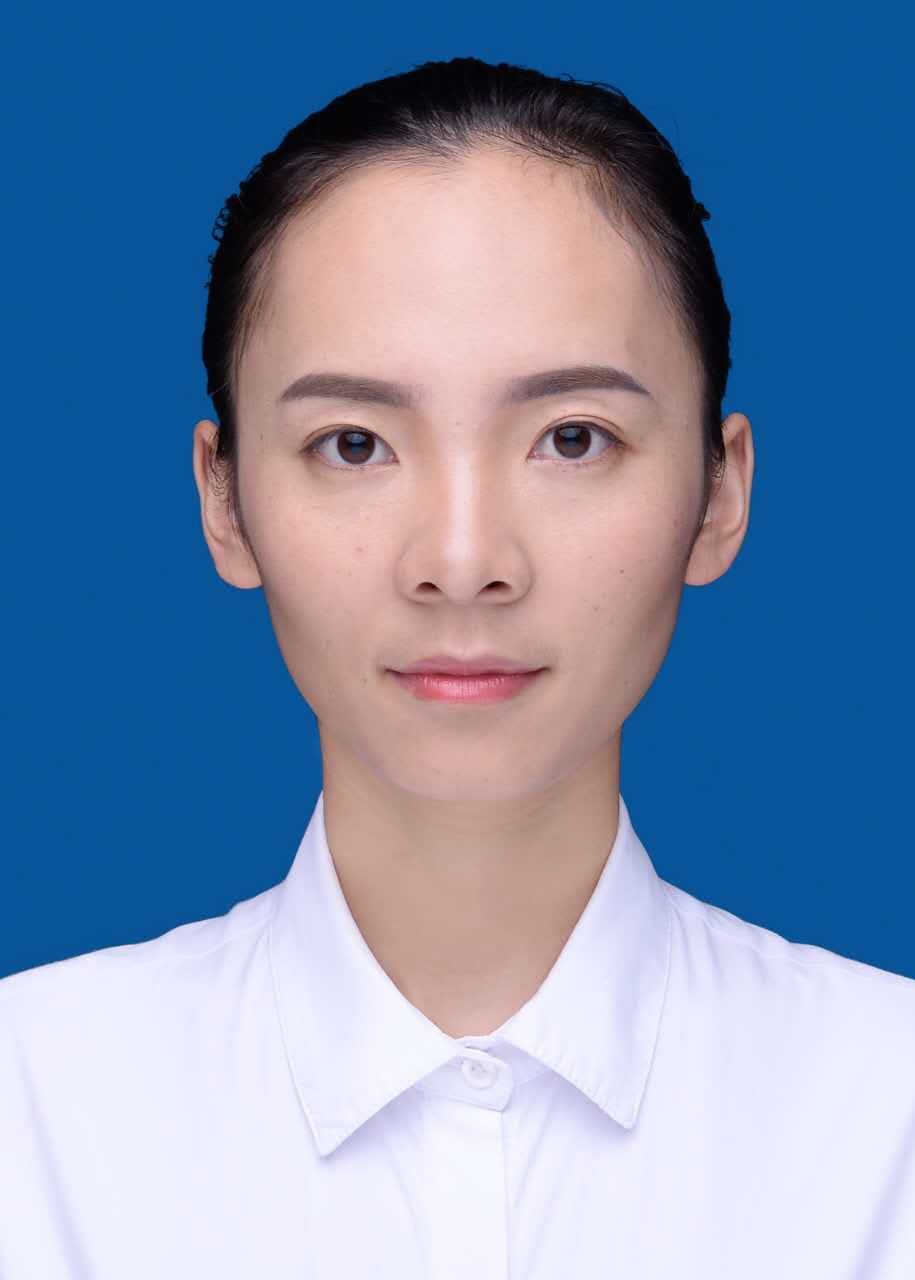}}]{Fuju Yan} received a bachelor's degree from Shandong University of Science and Technology in 2015, and a master's degree from China Jiliang University in 2018. She is currently pursuing the Ph.D. degree with School of Automation Engineering, Nanjing University of Aeronautics and Astronautics. Her research interests are image processing and target detection.
\end{IEEEbiography}



\vfill

\end{document}